\documentclass[10pt,twocolumn,letterpaper]{article}

\usepackage[pagenumbers]{cvpr} %

\usepackage[dvipsnames]{xcolor}
\usepackage{tikz}
\usetikzlibrary{positioning,fit,arrows.meta}
\usepackage{multirow}

\usepackage{colortbl}
\usepackage{arydshln}

\renewcommand{\paragraph}[1]{\vspace{0.2mm}\noindent\textbf{#1}}

\definecolor{cvprblue}{rgb}{0.21,0.49,0.74}
\usepackage[pagebackref,breaklinks,colorlinks,citecolor=cvprblue]{hyperref}

\def\paperID{85} %
\def\confName{3DV\xspace}
\def\confYear{2027\xspace}

\title{Smooth Essential-Manifold Residuals for View-Graph Calibration}

\author{Viktor Larsson\\
Lund University\\
\and
Johannes Schönberger\\
Meta Reality Labs\\
}

\begin{document}
\maketitle
\begin{abstract}
View-graph calibration jointly estimates camera intrinsics from multiple pairwise fundamental matrices.
The correct intrinsics transform each fundamental matrix into an essential matrix, casting the calibration problem as minimizing the distance to the essential manifold.

We study how to represent this objective for nonlinear least squares.
The conventional singular-value formulation collapses two independent essential-matrix constraints into one scalar residual.
Its per-edge Gauss--Newton model is therefore rank deficient, which leads to poor convergence.

We instead derive an isometric reduction to $2\times2$ matrices, where each branch of the nonlinear essential-matrix constraint becomes a linear subspace.
This reduction yields a smooth two-dimensional residual whose squared norm is exactly the original manifold distance, whose Gauss--Newton approximation is generically full rank, and whose evaluation requires no per-edge singular value decomposition.
Experiments show that the new residual retains the estimation accuracy of the geometric objective while converging in fewer, computationally cheaper iterations.
Combined with robust initialization and optimization, it achieves state-of-the-art view-graph calibration accuracy at a runtime comparable to inexpensive algebraic objectives.
\end{abstract}

\section{Introduction}
\label{sec:intro}

Camera intrinsics are required by Structure-from-Motion (SfM), visual localization, and multi-view stereo.
For internet photo collections and heterogeneous image sets, however, calibration metadata is often missing, incorrect, or invalidated by cropping and digital zoom.
At the same time, reconstruction systems estimate fundamental matrices between verified image pairs when building the view graph.
View-graph calibration uses these matrices to estimate intrinsics before recovering camera poses or 3D structure: each edge constrains both cameras by requiring its calibrated fundamental matrix to be essential.

Under common assumptions, a single fundamental matrix theoretically determines the focal lengths of both cameras through the closed-form solution of Bougnoux~\cite{bougnoux1998projective}.
In practice, two-view estimates are highly sensitive to image noise, weak or near-degenerate geometry, and outlier edges.
Joint estimation over the view graph is therefore required, as it lets each camera draw support from many pairwise constraints rather than depend on any single edge.

Existing view-graph calibration methods differ mainly in how they measure violation of the essential-matrix constraint, using polynomial invariants, normalized algebraic expressions, or Kruppa-derived residuals.
The normalized distance to the essential manifold is particularly appealing: it directly measures geometric inconsistency and is invariant to the arbitrary scale of the fundamental matrix and to scale changes induced by calibration~\cite{solonets2026global}.
Its conventional least-squares formulation, however, computes an SVD for every edge at every iteration and uses a scalar residual that is non-smooth at the solution, resulting in slow convergence.

\begin{figure}[t]
\centering
\includegraphics[width=\columnwidth]{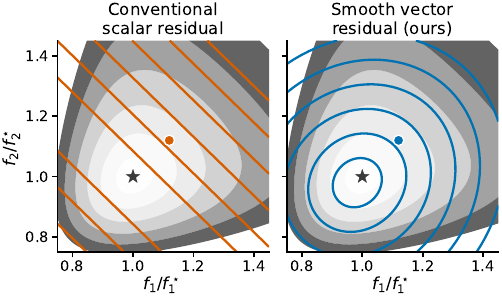}
\caption{
\textbf{Same objective, different local optimization models.}
Gray filled contours show the normalized essential-manifold cost for a single view-graph edge as its two focal lengths vary.
Colored contours show the Gauss--Newton approximation constructed at the marked iterate ({\Large$\bullet$}).
The conventional scalar residual produces a rank-one stripe model, whereas our smooth two-component factorization yields closed contours that better represent the local geometry of the same cost.
The star marks the solution.
}
\label{fig:teaser}
\end{figure}

We address these shortcomings without changing the objective.
By exploiting the rank-two structure of calibrated fundamental matrices, we obtain an equivalent $2\times2$ representation in which the essential constraint admits a smooth two-component residual.
Its squared norm is exactly the original normalized manifold distance, but its Jacobian yields a more informative local model and avoids repeated matrix decompositions.
The result is the same geometric cost in a form better suited to large-scale nonlinear least squares.

We incorporate the residual into a robust view-graph calibration pipeline that initializes focal lengths from a consensus of closed-form two-view estimates and solves the graph objective in two stages: a Cauchy warm-up followed by a Tukey loss that rejects outlier edges.
We evaluate it against algebraic, Kruppa-based, and SVD-based geometric formulations on standard benchmarks.
The smooth residual reaches the same minima as the SVD formulation in a fraction of the iterations and at a fraction of the cost per iteration, making the geometric objective as cheap to optimize as the algebraic ones.
In summary, our contributions are:
\begin{itemize}
\item We identify an isometric reduction of rank-two $3\times3$ matrices to $2\times2$ matrices that maps the essential manifold to the set of scaled orthogonal matrices.
\item This yields a smooth two-component residual whose squared norm is exactly the normalized manifold distance, giving a generically full-rank local model where an exact scalar residual is necessarily rank deficient and non-smooth at the solution.
\item We combine the residual with a consensus initialization and a two-stage robust schedule into a view-graph calibration pipeline.
\item We show experimentally that the reformulation substantially improves convergence and that the resulting pipeline is state of the art in accuracy while matching the runtime of the cheapest algebraic objectives.
\end{itemize}

\section{Related Work}
\label{sec:related_work}

\subsection{Self-Calibration from Epipolar Geometry}
\label{sec:related_selfcalib}

Classical self-calibration recovers camera intrinsics from the epipolar geometry of uncalibrated views.
The Kruppa equations \cite{kruppa1913ermittlung, faugeras1992camera, hartley1997kruppa} express the constraints imposed on the intrinsics by a fundamental matrix and form the basis of many two-view and multi-view methods.
Lourakis and Deriche~\cite{lourakis2000camera} express these constraints through the SVD of the fundamental matrix.
For two unknown focal lengths and known principal points, Bougnoux~\cite{bougnoux1998projective} gives a closed-form solution from a single fundamental matrix; algebraically equivalent direct and linear formulations were subsequently derived by Kanatani and Matsunaga~\cite{kanatani2000closedform} and Melanitis and Maragos~\cite{melanitis2021linear}.
Two-view focal recovery nevertheless remains sensitive to noise, weak geometry, and critical configurations such as intersecting optical axes~\cite{brooks1996recovering,sturm2001focal,sturm2005singular}.
Iterative methods improve robustness by incorporating approximate calibration or explicit priors~\cite{hartley2002approximate,kocur2024robust}.
This instability motivates aggregating focal-length constraints across multiple pairwise edges.

\subsection{View-Graph and Global Calibration}
\label{sec:related_viewgraph}

Multiview self-calibration can also recover a calibration shared across several views from image correspondences or compatible fundamental matrices.
Recent formulations avoid a full projective reconstruction by using point-depth constraints~\cite{dalcin2024minimal} or well-conditioned minimal relaxations of the three-view Kruppa system~\cite{kosaka2026minimal}.
Both approaches assume shared intrinsics. Kosaka et al.~\cite{kosaka2026general} extend Kruppa-based self-calibration to varying intrinsics, studying recoverability from a few views. We focus on optimizing the essential-manifold distance across a view graph.

Within this view-graph setting, methods couple the intrinsics of neighboring images through essential-matrix consistency and differ mainly in how they measure that consistency.
Sweeney et al.~\cite{sweeney2015optimizing} use a polynomial essential-matrix invariant within a broader viewing-graph optimization pipeline.
Geppert et al.~\cite{geppert2021privacy} normalize related algebraic invariants to remove the arbitrary scale of each edge.
Fetzer et al.~\cite{fetzer2020stable} derive focal-space residuals from the Kruppa equations, designed for stable optimization.
Mendon\c{c}a and Cipolla~\cite{mendoncca1999simple} minimize the relative gap $(\sigma_1-\sigma_2)/\sigma_2$ between the non-zero singular values.
Solonets et al.~\cite{solonets2026global} use the normalized distance to the essential manifold in a global focal-length and radial-distortion averaging framework.
In \cref{sec:objectives}, we discuss these objectives in more detail.

Solonets et al. are the closest to our work because we start from the same geometric and scale-invariant objective.
Rather than proposing another consistency loss or calibration model, we study how to represent this objective for nonlinear least squares.
Its conventional scalar residual requires repeated singular-value decompositions, is non-smooth at the solution, and gives a poor Hessian approximation in Gauss--Newton-style optimization.
We show that the same manifold distance admits an exact smooth residual representation that avoids these limitations.
The objective is unchanged; the difference lies in the residuals and the resulting optimization behavior.

\subsection{Calibration from Learned Priors}
\label{sec:related_learned}

Learned single-image methods infer intrinsics from learned monocular cues rather than multi-view geometry.
Recent examples are GeoCalib~\cite{veicht2024geocalib} and AnyCalib~\cite{tirado2025anycalib}.
Feed-forward reconstruction systems such as VGGT~\cite{wang2025vggt} and its successors~\cite{wang2026vggt,keetha2026mapanything} instead predict cameras together with scene geometry from an image collection.
They solve a broader task and rely on learned multi-image priors rather than a precomputed fundamental-matrix graph.
We include both families in our evaluation.

\section{Method}
\label{sec:method}

\subsection{View-Graph Calibration Problem}
\label{sec:problem}

We consider a view graph $\mathcal{G}=(\mathcal{V},\mathcal{P})$ in which each vertex is an image and each edge $(i,j)\in\mathcal{P}$ stores a rank-two fundamental matrix $F_{ij}$ estimated from feature matches.

Our goal is to estimate one focal length $f_i>0$ for each image.
We assume that the principal point $(u_i,v_i)$ is fixed to the image center, so the calibration matrix is
\begin{equation}
K_i(f_i)=
\begin{bmatrix}
f_i & 0 & u_i \\
0 & f_i & v_i \\
0 & 0 & 1
\end{bmatrix}.
\end{equation}
The calibrated fundamental matrix on an edge is
\begin{equation}
    A_{ij}(f_i,f_j)=K_j(f_j)^\top F_{ij}K_i(f_i).
    \label{eq:calibrated_fundamental}
\end{equation}
As the calibration matrices are invertible, $A_{ij}$ has rank two for all non-zero focal lengths.
In the noise-free case, the correct focal lengths make the two non-zero singular values equal and therefore place $A_{ij}$ on the essential manifold
\begin{equation}
    \mathcal{E}=\{U\operatorname{diag}(\alpha,\alpha,0)V^\top:\alpha>0,\ U,V\in\mathrm{O}(3)\}.
\end{equation}
Given a per-edge consistency loss $\ell$ that measures deviation from $\mathcal{E}$, view-graph calibration can be formulated as the robust graph optimization problem
\begin{equation}
    \min_{\{f_i>0\}}
    \sum_{(i,j)\in\mathcal{P}}
    \rho\!\left(\ell\bigl(A_{ij}(f_i,f_j)\bigr)\right),
    \label{eq:viewgraph_objective}
\end{equation}
where $\rho$ is a robust loss that limits the influence of outlier edges.
Each term couples only the two focal lengths incident to its edge, so the normal equations inherit the same sparsity structure as the view graph itself.

\subsection{Essential-Consistency Losses}
\label{sec:objectives}

Several losses have been proposed to measure essential-matrix consistency in \cref{eq:viewgraph_objective}.
We denote the residual block of each baseline by $r_{\mathrm{method}}$; its unrobustified per-edge least-squares cost is $\|r_{\mathrm{method}}\|_2^2$ and replaces $\ell$ in the graph objective.

\paragraph{Polynomial Invariant.}
Sweeney et al.~\cite{sweeney2015optimizing} penalize an essential-matrix invariant, which in our common least-squares notation is the scalar residual
\begin{equation}
    r_{\mathrm{Sweeney}}(A)
    =\|AA^\top\|_F^2-\tfrac{1}{2}\|A\|_F^4.
    \label{eq:sweeney_residual}
\end{equation}
For rank-two $A$, expanding the norms in its two non-zero singular values gives
\begin{equation}
    r_{\mathrm{Sweeney}}(A)
    =\tfrac{1}{2}(\sigma_1^2-\sigma_2^2)^2,
\end{equation}
so the invariant is non-negative and vanishes exactly when $\sigma_1=\sigma_2$.
The invariant is polynomial and inexpensive, but homogeneous of degree four in $A$.
Since the focal lengths affect $\|A\|_F$, it introduces a shrinking bias in addition to its dependence on the arbitrary edge scale.

\paragraph{Normalized Algebraic Ratios.}
Geppert et al.~\cite{geppert2021privacy} remove the scale dependence through the ratio residuals
\begin{equation}
    r_{\mathrm{Geppert}}(A)
    =
    \begin{bmatrix}
        \|AA^\top\|_F^2/\|A\|_F^4-1/2 \\[2pt]
        \|A\|_F^4/\|AA^\top\|_F^2-2
    \end{bmatrix}.
    \label{eq:geppert_residual}
\end{equation}
For rank-two $A$, the two components are respectively
\begin{equation}
    \frac{(\sigma_1^2-\sigma_2^2)^2}
         {2(\sigma_1^2+\sigma_2^2)^2},
    \qquad
    -\frac{(\sigma_1^2-\sigma_2^2)^2}
          {\sigma_1^4+\sigma_2^4},
\end{equation}
and therefore vanish exactly when $\sigma_1=\sigma_2$.
The ratios are scale invariant but have no direct geometric interpretation.

\paragraph{Kruppa-Derived Residuals.}
Fetzer et al.~\cite{fetzer2020stable} start from the Kruppa constraints of the principal-point-rectified fundamental matrix.
Each of the two constraints expresses one squared focal length as a rational function of the other, $\mathcal{K}_1(f_j^2)$ and $\mathcal{K}_2(f_i^2)$, with coefficients precomputed from the SVD of the rectified $F_{ij}$.
The residuals are the relative discrepancies
\begin{equation}
    r_{\mathrm{Fetzer}}(f_i,f_j)
    =
    \begin{bmatrix}
        \bigl(f_i^2-\mathcal{K}_1(f_j^2)\bigr)/f_i^2 \\[2pt]
        \bigl(f_j^2-\mathcal{K}_2(f_i^2)\bigr)/f_j^2
    \end{bmatrix},
    \label{eq:fetzer_residual}
\end{equation}
a form chosen to avoid the poles and poor convergence regions of earlier Kruppa objectives.

\paragraph{Essential-Manifold Distance.}
Mendon\c{c}a and Cipolla~\cite{mendoncca1999simple} penalize the singular-value gap directly, and Solonets et al.~\cite{solonets2026global} use the normalized squared Frobenius distance to the scaled essential manifold,
\begin{equation}
    \ell(A)
    =\frac{\operatorname{dist}^2(A,\mathcal{E})}{\|A\|_F^2}
    =\frac{(\sigma_1-\sigma_2)^2}{2(\sigma_1^2+\sigma_2^2)},
    \label{eq:normalized_manifold_distance}
\end{equation}
where $\sigma_1\geq\sigma_2$ are the non-zero singular values of $A$.
The normalization removes both the arbitrary scale of $F_{ij}$ and the scale changes induced by the focal lengths.

\begin{table}[t]
\centering
\caption{
\textbf{Essential-consistency objectives.}
``Scale inv.,'' ``Geom.,'' and ``Res. dim.'' denote scale invariance, geometric interpretation, and residual dimension.
``Gauss--Newton (GN) rank'' is the rank of the per-edge Hessian approximation $J^\top J$ where the residual is differentiable, which is what the solver uses as its local model; it falls below the residual dimension whenever the components are dependent.
``SVD'' marks objectives whose standard evaluation requires an SVD per edge inside the optimization loop.
}
\label{tab:loss_zoo}
\small
\setlength{\tabcolsep}{3pt}
\begin{tabular}{lccccc}
\hline
Objective & Scale inv. & Geom. & Res.\ dim. & GN rank & SVD \\
\hline
Sweeney~\eqref{eq:sweeney_residual} & $\times$ & $\times$ & 1 & 1 & $\times$ \\
Geppert~\eqref{eq:geppert_residual} & $\checkmark$ & $\times$ & 2 & 1 & $\times$ \\
Fetzer~\eqref{eq:fetzer_residual} & $\checkmark$ & $\times$ & 2 & 2 & $\times$ \\
Manifold~\eqref{eq:normalized_manifold_distance} & $\checkmark$ & $\checkmark$ & 1 & 1 & $\checkmark$ \\
Ours~($2\times 2$)~\eqref{eq:two_by_two_residual} & $\checkmark$ & $\checkmark$ & 2 & 2 & $\times$ \\
\hline
\end{tabular}
\end{table}

All of these objectives vanish exactly on $\mathcal{E}$, but they quantify deviations from it differently.
Only the manifold objective is a distance: in units of $\|A\|_F$, it measures how far an edge is from being essential rather than evaluating an algebraic invariant without a direct geometric interpretation.
Solonets et al.~\cite{solonets2026global} found this distance to yield the most accurate focal lengths among the objectives they evaluated, and our experiments confirm that result (\cref{sec:experiments}).
We therefore adopt \cref{eq:normalized_manifold_distance} as the starting objective and study how to represent it for least-squares optimization.

\subsection{Issues with the Manifold Distance}
\label{sec:standard_residual}

The conventional least-squares representation of \cref{eq:normalized_manifold_distance} has two limitations: its scalar factorization gives Gauss--Newton a poor local model, and its evaluation requires repeated singular value decompositions.
Nonlinear least squares writes each edge cost as the squared norm of a residual vector, linearizes that vector with Jacobian $J$, and uses $J^\top J$ to approximate the Hessian.
Different residual vectors can have the same squared norm $\ell=\|r\|^2$ yet induce different Gauss--Newton approximations $J^\top J$.

\paragraph{Two independent manifold constraints.}
Up to scale, the rank-two fundamental matrices form a seven-dimensional manifold, of which the essential matrices form a five-dimensional submanifold.
Membership in $\mathcal{E}$ is therefore a codimension-two condition: it constrains an edge by two independent scalar equations.

The conventional representation nevertheless compresses the two constraints into the single residual
\begin{equation}
    r_{\mathrm{manifold}}(A)
    =\sqrt{\ell(A)}
    =\frac{\sigma_1-\sigma_2}{\sqrt{2(\sigma_1^2+\sigma_2^2)}}.
    \label{eq:scalar_manifold_residual}
\end{equation}
Each edge depends on two focal lengths, but \cref{eq:scalar_manifold_residual} has a one-row Jacobian wherever it is differentiable, so its Gauss--Newton model has rank at most one and produces the stripe model in \cref{fig:teaser}.
One of the two directions of curvature is invisible to the solver.
This is a property of any one-dimensional residual rather than of this particular choice.

\begin{figure*}[t]
\centering
\includegraphics[width=\textwidth]{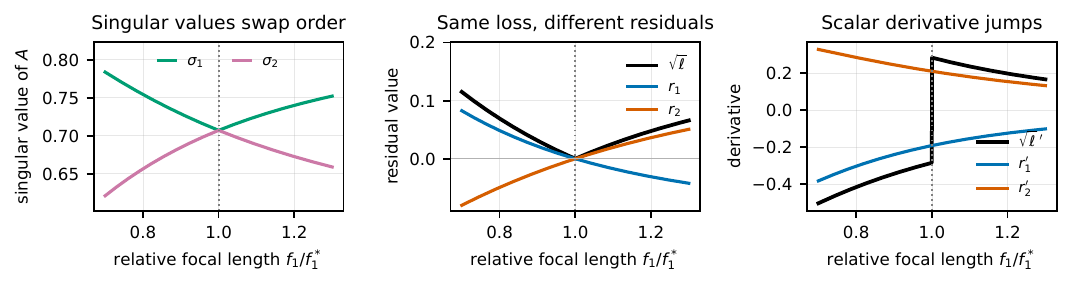}
\caption{
\textbf{Why the scalar residual kinks at the solution?}
Along a one-dimensional focal slice, the two non-zero singular values of $A_{ij}$ meet and exchange their ordering (left), so each turns a corner and their gap $\sqrt{\ell}=(\sigma_1-\sigma_2)/\sqrt{2}$ behaves like an absolute value.
The scalar residual therefore kinks exactly where the two signed $2\times2$ components vanish (middle) and its derivative jumps (right), while the components themselves cross zero smoothly with continuous derivatives.
}
\label{fig:em_kink}
\end{figure*}

The singular-value formulation also introduces a non-differentiability when the two singular-value branches cross, their ordered gap $\sigma_1-\sigma_2$ behaves like an absolute value, so its derivative jumps (\cref{fig:em_kink}).

\paragraph{Computational cost.}
The second limitation is evaluation cost.
Computing \cref{eq:normalized_manifold_distance} from its definition requires the singular values of every $A_{ij}$, and derivative evaluation must differentiate through the decomposition.
These repeated per-edge decompositions dominate the iteration cost on large graphs.
Removing the SVD alone would make each iteration cheaper but leave the scalar residual and its local model unchanged.
We therefore seek an exact representation that resolves the two constraints into separate differentiable components and avoids per-iteration SVDs.

\subsection{SVD-Free Manifold Distance}
\label{sec:three_by_three_loss}

The evaluation-cost problem can be solved on its own.
Similar to the costs of Sweeney et al.~\cite{sweeney2015optimizing} and Geppert et al.~\cite{geppert2021privacy}, the manifold distance can be reformulated directly in terms of the matrix $A$.
Since $F$ has rank two and the calibration matrices are invertible, $A$ has rank two by construction and $\sigma_3=0$.
Its two non-zero singular values therefore satisfy
\begin{equation}
    \begin{aligned}
    \|A\|_F^4-\|AA^\top\|_F^2
    &=(\sigma_1^2+\sigma_2^2)^2-(\sigma_1^4+\sigma_2^4) \\
    &=2\sigma_1^2\sigma_2^2.
    \end{aligned}
    \label{eq:singular_product_invariants}
\end{equation}
Combining \cref{eq:singular_product_invariants} with \cref{eq:normalized_manifold_distance} gives
\begin{equation}
    \ell(A)
    =\frac{1}{2}
     -\frac{\sqrt{\tfrac{1}{2}\bigl(\|A\|_F^4-\|AA^\top\|_F^2\bigr)}}
            {\|A\|_F^2}.
    \label{eq:svd_free_distance}
\end{equation}
Equation~\eqref{eq:svd_free_distance} evaluates the normalized manifold distance exactly using only matrix products and norms.

This removes the SVD from the loop, but nothing else changes: the objective is the same scalar, the residual $\sqrt{\ell}$ retains its kink, and the solver follows the same iterates at a lower cost per iteration.
Our experiments confirm this separation (\cref{sec:exp_ablation}): the SVD-free scalar reduces the time per iteration, while the iteration count is unchanged.

\subsection{Isometric Reduction to $2\times2$}
\label{sec:two_by_two_reduction}

The key structural observation is that the essential-matrix constraint simplifies when the rank-two matrix is expressed on the complements of its null spaces.
Let $a$ and $b$ be unit left and right null vectors of $A$, so that $A^\top a=0$ and $Ab=0$, and let $P_a,P_b\in\mathbb{R}^{2\times3}$ have orthonormal rows spanning $a^\perp$ and $b^\perp$.
Define
\begin{equation}
    C=P_aAP_b^\top\in\mathbb{R}^{2\times2}.
    \label{eq:two_by_two_reduction}
\end{equation}
Then $A=P_a^\top C P_b$, so the reduction loses no component of $A$, and the map $D\mapsto P_a^\top D P_b$ is a Frobenius isometry onto the subspace of matrices with these null spaces.
Under this identification the essential-matrix constraint becomes membership in the set of scaled orthogonal matrices:
\begin{equation}
    A\in\mathcal{E}
    \quad\Longleftrightarrow\quad
    C\in\{\alpha R:\alpha\geq0,\ R\in\mathrm{O}(2)\}.
    \label{eq:constraint_equivalence}
\end{equation}
\begin{samepage}
Scaled $\mathrm{O}(2)$ is the union of the scaled rotations $\mathcal{S}_+$ and the scaled reflections $\mathcal{S}_-$, which are orthogonal linear subspaces.
\begin{equation}
    \begin{gathered}
    \mathcal{S}_+
    =\left\{\begin{bmatrix}\alpha&\beta\\-\beta&\alpha\end{bmatrix}\right\},
    \qquad
    \mathcal{S}_-
    =\left\{\begin{bmatrix}\alpha&\beta\\\beta&-\alpha\end{bmatrix}\right\}, \\
    \alpha,\beta\in\mathbb{R}.
    \end{gathered}
    \label{eq:scaled_o2_subspaces}
\end{equation}
\end{samepage}

Each of these subspaces is two-dimensional, so the constraint set again has codimension two, now embedded in the four-dimensional space of $2\times2$ matrices, matching the count of \cref{sec:standard_residual}.

\subsection{Distance to $2\times2$ Scaled Orthogonal Matrices}
\label{sec:two_by_two_distance}

The main idea is that the reduction preserves the distance to the constraint set exactly:
\begin{equation}
    \operatorname{dist}(A,\mathcal{E})
    =\operatorname{dist}\bigl(C,\mathcal{S}_+\cup\mathcal{S}_-\bigr).
    \label{eq:isometric_distance}
\end{equation}
This follows because (i) $C$ has the same rank and the same non-zero singular values as $A$, with $\|C\|_F=\|A\|_F$, and (ii) the closest essential matrix to $A$ can be chosen with the same null spaces as $A$, so restricting the projection to the isometric subspace does not change the distance.
Formal proofs are given in the supplement.

The distance to the nearer linear subspace can be expressed by the smooth two-component residual
\begin{equation}
    r_{ij}(f_i,f_j)
    =\frac{1}{\sqrt{2}\,\|A_{ij}\|_F}
    \begin{bmatrix}
        c_{11}-\eta c_{22}\\
        c_{12}+\eta c_{21}
    \end{bmatrix}.
    \label{eq:two_by_two_residual}
\end{equation}
where $C=\bigl[\begin{smallmatrix}c_{11}&c_{12}\\c_{21}&c_{22}\end{smallmatrix}\bigr]$ and $\eta=\operatorname{sign}(\det C)$ selects the nearer subspace.
The particular choice of projections is immaterial: any other pair replaces $C$ by $R_aCR_b^\top$ with $R_a,R_b\in\mathrm{O}(2)$, which maps $\mathcal{S}_+\cup\mathcal{S}_-$ onto itself and therefore leaves $\|r_{ij}\|$ and the per-edge cost unchanged.
By \cref{eq:isometric_distance},
\begin{equation}
    \|r_{ij}(f_i,f_j)\|_2^2=\ell(A_{ij}).
    \label{eq:residual_equivalence}
\end{equation}
For a non-degenerate edge this gives a full-rank $2\times 2$ Jacobian at the solution, so the per-edge Gauss--Newton Hessian $J^\top J$ is also full rank, allowing for a better approximation (see \cref{fig:teaser}).
\Cref{fig:em_kink} compares the two representations along a one-dimensional slice through the solution: $\sqrt{\ell}$ is the norm of the two signed components, so it kinks where they both vanish and its derivative jumps, while the components themselves cross zero smoothly.
The SVD-based distance, the SVD-free $3\times3$ scalar, and the $2\times2$ residual block therefore optimize exactly the same objective; they only differ in what approximate Hessian $J^\top J$ they produce.

\subsection{Parametric Projections and Branching}
\label{sec:smooth_evaluation}

The reduction requires the null spaces of $A_{ij}$, but these never have to be recomputed during optimization.
Let $a_{ij}$ and $b_{ij}$ be fixed left and right null vectors of $F_{ij}$, computed once when the edge is constructed.
The unit null vectors of the calibrated matrix follow by transport through the current calibration matrices,
\begin{equation}
    \hat{a}_{ij}=\frac{K_j^{-1}a_{ij}}{\|K_j^{-1}a_{ij}\|},
    \qquad
    \hat{b}_{ij}=\frac{K_i^{-1}b_{ij}}{\|K_i^{-1}b_{ij}\|}.
    \label{eq:calibrated_null_vectors}
\end{equation}
The complements $P_a$ and $P_b$ then follow in closed form.
Writing $a_{ij}=(a_1,a_2,a_3)^\top$ and $(u,v)$ for the principal point of image $j$,
\begin{equation}
    K_j(f_j)^{-1}a_{ij}
    \propto(a_1-ua_3,\ a_2-va_3,\ f_ja_3)^\top,
    \label{eq:null_vector_plane}
\end{equation}
so $\hat{a}_{ij}$ stays in the plane spanned by $e_3$ and $w=(a_1-ua_3,\,a_2-va_3,\,0)^\top$ for every focal length, and this plane is fixed because $w$ does not depend on $f_j$.
The unit vector $n_a\propto w\times e_3$ is therefore orthogonal to $\hat{a}_{ij}$ on the entire domain, and
\begin{equation}
    P_a=\begin{bmatrix}n_a^\top\\(\hat{a}_{ij}\times n_a)^\top\end{bmatrix}
    \label{eq:complement}
\end{equation}
has orthonormal rows spanning $\hat{a}_{ij}^\perp$.
Note that the two rows are orthogonal unit vectors, and $\|\hat{a}_{ij}\times n_a\|=1$ precisely because $\hat{a}_{ij}$ and $n_a$ are orthogonal and both of unit length, so no further normalization is needed.
The corresponding $n_b$ is obtained from $b_{ij}$ and $K_i$.
Both reference axes are computed once per edge, so \cref{eq:complement} varies smoothly with the focal lengths and maintains a constant orientation (see supplementary for more details).
Every per-iteration operation (calibration, normalization, the $2\times3$ complements, and the reduction \cref{eq:two_by_two_reduction}) is an elementary matrix expression that is cheap to evaluate and to differentiate, analytically or by automatic differentiation.

The branch $\eta$ in \cref{eq:two_by_two_residual} is likewise fixed once at initialization.
Since $|\det C|=\sigma_1\sigma_2>0$ for every positive focal pair and the positive focal domain is connected, the continuous function $\det C$ cannot change sign along any optimization trajectory.
With the reference axes and $\eta$ fixed, each component of \cref{eq:two_by_two_residual} is smooth in the focal lengths.

\subsection{Robust View-Graph Calibration Pipeline}
\label{sec:pipeline}

We use the proposed residual in a view-graph calibration pipeline optimizing \cref{eq:viewgraph_objective}.
We parameterize the log-focals to enforce positivity, $f_i = \exp(s_i)$.

\paragraph{Initialization.}
Focal lengths are initialized from the per-image median of the focal lengths from the closed-form two-view estimates~\cite{bougnoux1998projective}.
If the votes are insufficient or inconsistent, each image’s COLMAP prior is scaled by a scene-level median estimated-to-prior focal ratio.
Further details are provided in the supplementary material.
Outlier and near-degenerate edges produce erratic per-edge estimates, but the subsequent warm-up stage tolerates the resulting initialization errors.

\paragraph{Two-Stage Robust Optimization.}
Estimated view graphs contain a substantial fraction of outlier edges whose fundamental matrices are inconsistent with any focal configuration.
We therefore solve \cref{eq:viewgraph_objective} twice with different robust losses $\rho$.
The first solve uses a Cauchy loss at a fixed scale.
Its influence function never vanishes, so this warm-up reaches the correct basin even from inaccurate initializations.
The second solve starts from the warm-up solution and uses a redescending Tukey loss.
Edges with residuals beyond the Tukey cutoff receive exactly zero influence, making the final estimate insensitive to outlier edges without any explicit edge filtering or trimming.

\section{Experiments}
\label{sec:experiments}

\subsection{Experimental Setup}
\label{sec:exp_setup}

\begin{figure*}[t]
\centering
\footnotesize
\setlength{\tabcolsep}{3pt}
\newsavebox{\datasetstatisticstable}
\sbox{\datasetstatisticstable}{\begin{tabular}{lrrrrrr}
\toprule
Dataset & scenes & images & focals & edges & matches & bad@$10^\circ$ \\
\midrule
ETH3D & 25 & 898 & 31 & 8k & 5M & 0.01 \\
B-MVS & 113 & 17,819 & 126 & 403k & 120M & 0.10 \\
IMC23 & 22 & 5,788 & 3,412 & 904k & 273M & 0.17 \\
IMC24 & 7 & 2,117 & 397 & 23k & 10M & 0.14 \\
\midrule
all & 167 & 26,622 & 3,966 & 1.34M & 408M & 0.09 \\
\bottomrule
\end{tabular}
}
\newlength{\datasetstatisticsplotwidth}
\setlength{\datasetstatisticsplotwidth}{\dimexpr\linewidth-\wd\datasetstatisticstable-6pt\relax}
\begin{minipage}[t]{\wd\datasetstatisticstable}
\vspace{0pt}
\usebox{\datasetstatisticstable}
\end{minipage}\hfill
\begin{minipage}[t]{\datasetstatisticsplotwidth}
\vspace{0pt}
\vspace{-0.5cm}
\includegraphics[width=0.9\linewidth]{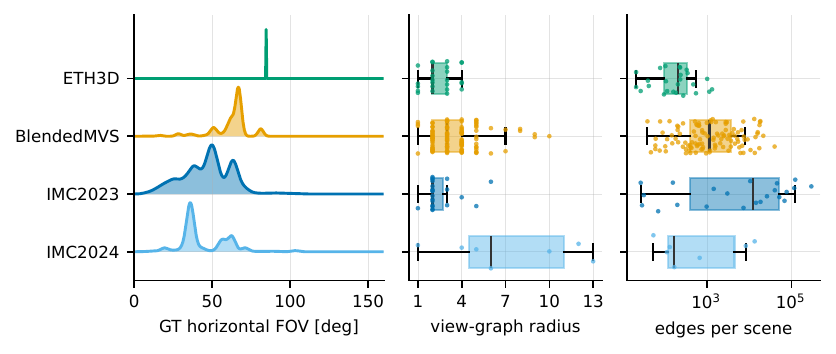}
\end{minipage}
\caption{
\textbf{Benchmark overview.}
The table counts scenes, ground-truth-covered images, unique focal lengths, view-graph edges after the minimum-match filter, and verified matches.
bad@$10^\circ$ is the median per-scene fraction of edges whose relative rotation, after upgrading the estimated $F$ with the ground-truth intrinsics, deviates more than $10^\circ$ from the ground-truth relative pose.
}
\label{fig:dataset_statistics}
\end{figure*}

\paragraph{Datasets and Ground Truth.}
We evaluate on ETH3D~\cite{schops2017multi}, BlendedMVS (B-MVS)~\cite{yao2020blendedmvs}, IMC2023~\cite{imc2023}, IMC2024~\cite{imc2024}, and IMC2025~\cite{imc2025}, using reference intrinsics from the released calibrations or SfM models as ground truth (GT).
\Cref{fig:dataset_statistics} summarizes the benchmark: 167 scenes with 26{,}622 ground-truth-covered images and 1.33M view-graph edges, spanning single-lens and per-image-focal regimes, fields of view from $4^\circ$ to $156^\circ$, and both hub-like and chain-like graph topologies.
As it lacks ground-truth focal lengths, IMC2025~\cite{imc2025} is only used for the pose evaluation.

\paragraph{Graph Construction and Baselines.}
We use COLMAP~\cite{schonberger2016structure} for feature extraction, matching, and geometric verification, and PoseLib~\cite{larsson2020poselib} to refine the fundamental matrices.
Edges with fewer than 50 verified matches are discarded.
The main evaluation is against the different loss functions listed in \cref{sec:objectives}.
For comparison, we also include the off-the-shelf COLMAP view-graph calibrator, which uses the loss of Fetzer et al.~\cite{fetzer2020stable}.

\paragraph{Metrics.}
The focal error of an image is the relative error $|\hat f-f^\star|/f^\star$ against the ground-truth focal length.
We report the median relative error and the area under curve (AUC) at 5\% and 20\%, as fractions in $[0,1]$.
Per-image errors are pooled across the scenes of a dataset rather than averaged per scene, so large scenes carry proportionally more weight.
The median is taken over the images that received an estimate, while the AUC is computed over all images.

\paragraph{Implementation.}
The global optimization is implemented in Ceres~\cite{ceres} and parametrized in log-focal space.
Unless stated otherwise, all methods use the same view graphs, the same Bougnoux-consensus initialization, and the identical two-stage robust schedule of \cref{sec:pipeline}: a Cauchy warm-up with a shared scale of $10^{-3}$, followed by a warm-started Tukey solve whose cutoff is the one tuned constant per method ($0.01$ for the essential-manifold residuals, a scale on $\|r\|$).
The parameters are tuned independently for each method on IMC2023.
Principal points are fixed to the image centers for all methods.

\subsection{Main Results}
\label{sec:exp_main}

\Cref{tab:main_results} reports focal accuracy against all baselines per dataset, as well as the end-to-end calibration time.

\begin{table*}[t]
\centering
\caption{
\textbf{Focal-length calibration results per dataset.}
Median relative focal error in percent (lower is better), AUC of the relative-error curve at 5\% and 20\% (higher is better), and mean end-to-end calibration time per scene in seconds.
All methods share the same view graphs; all but the COLMAP calibrator, which runs as shipped, also share the initialization and robust schedule (\cref{sec:exp_setup}).
Bold and underline mark the best and second-best value per column.
}
\label{tab:main_results}
\footnotesize
\setlength{\tabcolsep}{3pt}
\setlength{\tabcolsep}{1.4pt}
\begin{tabular}{lc*{4}{c}c*{4}{c}c*{4}{c}c*{4}{c}}
\toprule
& & \multicolumn{4}{c}{ETH3D} & & \multicolumn{4}{c}{BlendedMVS} & & \multicolumn{4}{c}{IMC2023} & & \multicolumn{4}{c}{IMC2024} \\
\cmidrule(lr){3-6} \cmidrule(lr){8-11} \cmidrule(lr){13-16} \cmidrule(lr){18-21}
Method & & med. & @5\% & @20\% & time & & med. & @5\% & @20\% & time & & med. & @5\% & @20\% & time & & med. & @5\% & @20\% & time \\
\midrule
COLMAP VGC & & 0.71 & 0.632 & 0.751 & 0.22 & & 4.90 & 0.383 & 0.541 & 0.52 & & 3.83 & 0.340 & 0.647 & \textbf{2.76} & & 8.96 & 0.225 & 0.468 & 1.23 \\
\arrayrulecolor{gray}\hdashline\arrayrulecolor{black}
\noalign{\vskip 1pt}
Sweeney et al. & & 0.61 & 0.752 & 0.856 & \underline{0.05} & & 99.98 & 0.242 & 0.286 & \textbf{0.39} & & 100.00 & 0.015 & 0.020 & \underline{3.47} & & 99.99 & 0.181 & 0.282 & 0.75 \\
Geppert et al. & & \underline{0.35} & 0.855 & 0.943 & \underline{0.05} & & \underline{0.63} & 0.732 & 0.860 & 0.53 & & 1.27 & 0.596 & \underline{0.807} & 6.07 & & 2.65 & 0.423 & 0.694 & 0.90 \\
Fetzer et al. & & 0.68 & 0.742 & 0.905 & \textbf{0.04} & & 1.10 & 0.646 & 0.839 & \underline{0.40} & & 1.85 & 0.524 & 0.784 & 4.78 & & 3.84 & 0.335 & 0.643 & \textbf{0.38} \\
Ess. manifold (SVD) & & \textbf{0.24} & \underline{0.886} & \underline{0.964} & 0.10 & & \textbf{0.56} & \underline{0.762} & \underline{0.887} & 1.08 & & \underline{1.24} & \underline{0.608} & \textbf{0.818} & 15.47 & & \textbf{2.46} & \textbf{0.452} & \textbf{0.738} & 1.59 \\
Ours & & \textbf{0.24} & \textbf{0.894} & \textbf{0.966} & \textbf{0.04} & & \textbf{0.56} & \textbf{0.769} & \textbf{0.896} & 0.43 & & \textbf{1.22} & \textbf{0.611} & \textbf{0.818} & 5.37 & & \underline{2.63} & \underline{0.435} & \underline{0.722} & \underline{0.46} \\
\bottomrule
\end{tabular}

\end{table*}

Across all four datasets, the two essential-manifold formulations attain the highest AUC at both thresholds, indicating that the geometric distance is a more effective calibration objective than the algebraic invariants.
Geppert et al.~\cite{geppert2021privacy} is the strongest algebraic baseline but consistently attains lower AUC scores across the board.
On every dataset, our residual and robust pipeline outperform the default COLMAP view-graph calibrator.

Our runtime is comparable to that of the algebraic view-graph baselines; \cref{sec:exp_ablation} analyzes the cost of evaluating the essential-manifold formulations.

\subsection{Ablation Studies}
\label{sec:exp_ablation}

\begin{figure}[t]
\centering
\includegraphics[width=\columnwidth]{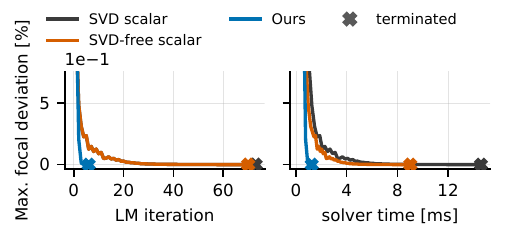}
\caption{
\textbf{Convergence on the ETH3D Playground scene} (38 images, 226 edges), for the SVD scalar residual, its SVD-free $3\times3$ equivalent, and the smooth $2\times2$ residual.
All methods optimize the Tukey objective of \cref{eq:viewgraph_objective} directly from the same consensus initialization.
The vertical axis is the maximum camera-wise relative deviation from each run's terminal focal vector, and crosses mark termination.
The common initial iterate and first update enter from above the displayed linear range.
}
\label{fig:em_convergence}
\end{figure}

\paragraph{Convergence and Runtime.}
We first compare the three residual representations that optimize the same objective: the original scalar SVD residual of \cref{eq:scalar_manifold_residual}, the SVD-free $3\times3$ scalar of \cref{eq:svd_free_distance}, and the smooth $2\times2$ residual of \cref{eq:two_by_two_residual}.
\Cref{fig:em_convergence} shows a qualitative example of the convergence for the ETH3D Playground scene.
The proposed 2-dimensional residual yields significantly faster convergence by providing better approximate Hessians to the non-linear least squares solver.

Next, in \cref{tab:essential_cost_evaluation} we compare the computational cost, both for evaluating the residual, but also for the Jacobian computation.
The proposed formulation is significantly cheaper to compute compared to the SVD-based version.

\begin{table}[t]
\centering
\caption{
\textbf{Per-edge evaluation cost} of the three equivalent essential-distance formulations, measured in a single-thread on an Intel Core i7-12700KF.
Each cell reports median nanoseconds per edge and speedup relative to the SVD formulation in the same row.
The Jacobian row includes residual evaluation, as required by the Ceres interface.
One-time construction, including the null-space SVD of the $2\times2$ formulation, is excluded.
}
\label{tab:essential_cost_evaluation}
\footnotesize
\setlength{\tabcolsep}{3pt}
\resizebox{\columnwidth}{!}{
\begin{tabular}{lccc}
\toprule
Evaluation & SVD scalar & No-SVD scalar & $2\times2$ \\
\midrule
Residual & 513 ns ($1.00\times$) & 57 ns ($8.98\times$) & \textbf{47 ns ($10.91\times$)} \\
Residual + Jacobian & 548 ns ($1.00\times$) & 112 ns ($4.89\times$) & \textbf{81 ns ($6.79\times$)} \\
\bottomrule
\end{tabular}

}
\end{table}

\paragraph{Robust Schedule and Initialization.}
We jointly ablate the initialization and robust schedule of our $2\times2$ formulation in \cref{tab:pipeline_ablation}.
The combination of consensus initialization and two-stage optimization is either best or close to the best configuration on every dataset, and is best on average.

\begin{table}[t]
\centering
\caption{
\textbf{Robust-schedule and initialization ablation.}
AUC of the relative focal error at 5\% for our $2\times2$ residual, either with prior initialization or consensus, combined with the three robust schedules.
Bold and underline mark the best and second-best configuration per column.
Proposed method highlighted in gray.
}
\label{tab:pipeline_ablation}
\footnotesize
\setlength{\tabcolsep}{3pt}
\begin{tabular}{lrrrrr}
\toprule
Schedule & ETH3D & B-MVS & IMC23 & IMC24 & Avg. \\
\midrule
\multicolumn{6}{l}{\emph{Prior initialization}} \\
\quad Cauchy only & \underline{0.895} & 0.746 & 0.536 & \underline{0.440} & 0.654 \\
\quad Tukey only & 0.390 & 0.610 & 0.527 & 0.205 & 0.433 \\
\quad Two-stage & 0.890 & \textbf{0.769} & \underline{0.609} & 0.425 & \underline{0.673} \\
\addlinespace
\multicolumn{6}{l}{\emph{Consensus initialization}} \\
\quad Cauchy only & \textbf{0.899} & 0.746 & 0.550 & \textbf{0.448} & 0.661 \\
\quad Tukey only & 0.888 & \underline{0.754} & 0.539 & 0.336 & 0.629 \\
\rowcolor{gray!25} \quad Two-stage \textbf{(default)} & 0.894 & \textbf{0.769} & \textbf{0.611} & 0.435 & \textbf{0.677} \\
\bottomrule
\end{tabular}

\end{table}

\subsection{Comparison to Learned Calibration}
\label{sec:exp_learned}

We compare our view-graph calibration (VGC) to the learned single-image calibrators GeoCalib~\cite{veicht2024geocalib} and AnyCalib~\cite{tirado2025anycalib}, and to the feed-forward reconstruction models VGGT-$\Omega$~\cite{wang2026vggt} and MapAnything~\cite{keetha2026mapanything}.
The comparison uses the ETH3D, IMC2023, and IMC2024 datasets of the main focal benchmark, comprising 8{,}803 cameras.
We run the released GeoCalib and AnyCalib pinhole models independently on every reference image.
For MapAnything, we use the released v1.1 research checkpoint with memory-efficient inference.
For VGGT-$\Omega$ and MapAnything, we sort each scene's input images by name and process them in deterministic, non-overlapping chunks of at most 200 views; chunks are inferred independently.
All learned methods return separate horizontal and vertical focal lengths, which we reduce to the scalar $\hat f=(\hat f_x+\hat f_y)/2$ used by our benchmark.
For the COLMAP reconstruction, we use the automatic pipeline's high-quality setting, with stricter feature extraction and geometric verification than in the focal-length evaluation; the calibration results therefore differ slightly from \cref{tab:main_results}.

\Cref{tab:learned_calibration} shows that AnyCalib consistently improves on GeoCalib for focal estimation.
VGGT-$\Omega$ is the strongest learned baseline throughout, while MapAnything improves substantially over the single-image calibrators on ETH3D and IMC2023 but not on IMC2024.
While VGGT-$\Omega$ outperforms baseline COLMAP on some metrics, the proposed view-graph calibration pipeline is consistently the best.

\begin{table}[t]
\centering
\caption{
\textbf{Comparison to learned calibration.}
Focal accuracy on the exact image inventory and with the pooled per-image metrics of \cref{tab:main_results}.
The methods use different inputs and solve different tasks, as grouped by information regime.
Bold and underline mark the best and second-best method per dataset and metric.
}
\label{tab:learned_calibration}
\footnotesize
\setlength{\tabcolsep}{3pt}
\resizebox{\linewidth}{!}{
\begin{tabular}{lrrrrrrrrr}
\toprule
& \multicolumn{3}{c}{ETH3D} & \multicolumn{3}{c}{IMC23} & \multicolumn{3}{c}{IMC24} \\
 \cmidrule(lr){2-4} \cmidrule(lr){5-7} \cmidrule(lr){8-10}
Method & med. & @5\% & @20\% & med. & @5\% & @20\% & med. & @5\% & @20\% \\
\midrule
 GeoCalib & 7.81 & 0.203 & 0.527 & 10.23 & 0.148 & 0.452 & 29.63 & 0.055 & 0.194 \\
AnyCalib& 5.73 & 0.242 & 0.636 & 7.65 & 0.193 & 0.542 & 11.58 & 0.121 & 0.412 \\
\midrule
 VGGT-$\Omega$ & \underline{0.53} & \underline{0.813} & \underline{0.952} & \underline{1.68} & \underline{0.596} & \underline{0.863} & 8.76 & 0.134 & \underline{0.501} \\
MapAny. & 3.25 & 0.386 & 0.704 & 2.46 & 0.459 & 0.794 & 14.84 & 0.061 & 0.309 \\
\midrule
COLMAP & 0.78 & 0.618 & 0.762 & 2.17 & 0.485 & 0.790 & \underline{7.42} & \underline{0.229} & 0.490 \\
Ours & \textbf{0.27} & \textbf{0.882} & \textbf{0.964} & \textbf{1.03} & \textbf{0.677} & \textbf{0.881} & \textbf{2.67} & \textbf{0.432} & \textbf{0.728} \\
\bottomrule
\end{tabular}

}
\end{table}

\subsection{End-to-End Reconstruction}
\label{sec:exp_downstream}
Finally, we evaluate the impact of view-graph calibration on SfM.
For this experiment, we additionally include IMC2025~\cite{imc2025}, which contains distractor images and often produces multiple reconstruction components.
We run COLMAP~\cite{schonberger2016structure} and GLOMAP~\cite{pan2024global} based on the latest COLMAP 4.1.1 release with their native calibration paths.
For \textit{Ours} we run the view-graph calibration first and provide the estimated focal lengths to the reconstruction pipeline.
\Cref{tab:downstream_reconstruction} reports pooled AUC at $5^\circ$ and $20^\circ$ for both the calibrated two-view poses estimated during reconstruction and for the final poses after end-to-end reconstruction.

\begin{table}[t]
\centering
\caption{
\textbf{Two-view geometry and downstream reconstruction pose accuracy.}
Two-view geometry (TVG) error is the maximum of rotation and translation-direction error for the estimated relative poses.
The end-to-end reconstruction errors compare reconstructed relative poses against the ground-truth.
}
\label{tab:downstream_reconstruction}
\footnotesize
\setlength{\tabcolsep}{4pt}
\begin{tabular}{llrrrrrr}
\toprule
& & \multicolumn{2}{c}{TVG} & \multicolumn{2}{c}{COLMAP} & \multicolumn{2}{c}{GLOMAP} \\
\cmidrule(lr){3-4} \cmidrule(lr){5-6} \cmidrule(lr){7-8}
Dataset & Calib. & @$5^\circ$ & @$20^\circ$ & @$5^\circ$ & @$20^\circ$ & @$5^\circ$ & @$20^\circ$ \\
\midrule
ETH3D & Original & \underline{0.640} & \underline{0.785} & \underline{0.847} & \underline{0.917} & \underline{0.875} & \underline{0.956} \\
 & Ours & \textbf{0.823} & \textbf{0.917} & \textbf{0.856} & \textbf{0.928} & \textbf{0.884} & \textbf{0.963} \\
\midrule
IMC2023 & Original & \underline{0.361} & \underline{0.630} & \underline{0.804} & \textbf{0.865} & \textbf{0.801} & \underline{0.860} \\
 & Ours & \textbf{0.428} & \textbf{0.672} & \textbf{0.805} & \underline{0.863} & \textbf{0.801} & \textbf{0.861} \\
\midrule
IMC2024 & Original & \underline{0.301} & \underline{0.616} & \underline{0.350} & \underline{0.394} & \underline{0.316} & \underline{0.395} \\
 & Ours & \textbf{0.501} & \textbf{0.779} & \textbf{0.353} & \textbf{0.395} & \textbf{0.325} & \textbf{0.412} \\
\midrule
IMC2025 & Original & \underline{0.272} & \underline{0.563} & \underline{0.379} & \textbf{0.442} & \underline{0.187} & \underline{0.221} \\
 & Ours & \textbf{0.379} & \textbf{0.652} & \textbf{0.382} & \textbf{0.442} & \textbf{0.191} & \textbf{0.222} \\
\bottomrule
\end{tabular}
\end{table}

Ours substantially improves the two-view geometries on every dataset, while its effect on end-to-end pose accuracy is smaller and occasionally neutral.
The gains are generally larger for global SfM, which depends more directly on the quality of the two-view geometries.

\section{Conclusion}
\label{sec:conclusions}

\paragraph{Limitations.}
View-graph calibration remains dependent on informative pairwise geometry: scenes whose edges are dominated by weak baselines or critical configurations constrain the focal lengths only weakly, and no reformulation of the objective can add information that the fundamental matrices do not carry.
Our main benchmark uses a single focal length per image with fixed principal points.
However, in principle the method allows for optimizing arbitrary intrinsic parameters.

\paragraph{Conclusion.}
We showed that the normalized essential-manifold distance, a geometrically meaningful and scale-invariant objective for view-graph calibration, admits an exact smooth two-component residual.
An isometric reduction of the rank-two calibrated fundamental matrix to $2\times2$ transforms the non-linear essential-matrix constraints into simple linear constraints.
The resulting residual preserves the original objective while giving Gauss--Newton and Levenberg--Marquardt a well-defined, generically full-rank local model at the solution, substantially improving convergence.
The geometric objective becomes competitive in runtime with inexpensive algebraic baselines while attaining better accuracy, making it a practical default for calibrating large view graphs before reconstruction.

\paragraph{Acknowledgements.}
This work was supported by the Wallenberg Artificial
Intelligence, Autonomous Systems and Software Program
(WASP), funded by the Knut and Alice Wallenberg Foundation, the Crafoord Foundation, and by the strategic research environment ELLIIT, funded by the Swedish government.

{
    \small
    \bibliographystyle{ieeenat_fullname}
    \bibliography{main}
}
\clearpage
\setcounter{page}{1}
\maketitlesupplementary

\section{Proofs for the Isometric Reduction}
\label{sec:suppl_proofs}

This section proves the claims of \cref{sec:two_by_two_reduction,sec:two_by_two_distance,sec:smooth_evaluation}.
Throughout, $A\in\mathbb{R}^{3\times3}$ has rank two with singular values $\sigma_1\geq\sigma_2>\sigma_3=0$, unit left and right null vectors $a$ and $b$, and $P_a,P_b\in\mathbb{R}^{2\times3}$ have orthonormal rows spanning $a^\perp$ and $b^\perp$.

\subsection{Frobenius Isometry of the Reduction}
\label{sec:suppl_isometry}

Let $Q_a=\bigl[\begin{smallmatrix}P_a\\a^\top\end{smallmatrix}\bigr]$ and $Q_b=\bigl[\begin{smallmatrix}P_b\\b^\top\end{smallmatrix}\bigr]$, which are orthogonal $3\times3$ matrices by construction.
Since $A^\top a=0$ and $Ab=0$,
\begin{equation}
    Q_aAQ_b^\top
    =
    \begin{bmatrix}
        P_aAP_b^\top & P_aAb\\
        a^\top AP_b^\top & a^\top Ab
    \end{bmatrix}
    =
    \begin{bmatrix}
        C & 0\\
        0 & 0
    \end{bmatrix}.
\end{equation}
Orthogonal matrices preserve singular values and Frobenius norms, so $C$ has the same rank and non-zero singular values as $A$, $\|C\|_F=\|A\|_F$, and $A=P_a^\top CP_b$.
For any $D\in\mathbb{R}^{2\times2}$, $\|P_a^\top DP_b\|_F=\|D\|_F$ by the same argument, so $D\mapsto P_a^\top DP_b$ is a Frobenius isometry onto the subspace of matrices with left null vector $a$ and right null vector $b$.

\subsection{Closest Essential Matrix Shares the Null Spaces}
\label{sec:suppl_nullspace}

Let $A=U\operatorname{diag}(\sigma_1,\sigma_2,0)V^\top$ be an SVD.
By orthogonal invariance of the Frobenius norm, its projection onto the scaled essential manifold is $\hat{E}=U\operatorname{diag}(\bar\sigma,\bar\sigma,0)V^\top$, where minimizing $(\sigma_1-\alpha)^2+(\sigma_2-\alpha)^2$ gives $\bar\sigma=\tfrac{1}{2}(\sigma_1+\sigma_2)$.
Its left and right null vectors are the third columns of $U$ and $V$, which are exactly the null vectors $a$ and $b$ of the rank-two matrix $A$.
Hence the projection stays inside the isometric subspace of \cref{sec:suppl_isometry}, and restricting the distance computation to that subspace does not change the minimum:
$\operatorname{dist}(A,\mathcal{E})=\operatorname{dist}(C,\{\alpha R:\alpha\geq0,\ R\in\mathrm{O}(2)\})$.

\subsection{Scaled \texorpdfstring{$\mathrm{O}(2)$}{O(2)} Is a Union of Two Orthogonal Subspaces}
\label{sec:suppl_o2}

Every scaled rotation and scaled reflection can be written as
\begin{equation}
    \begin{bmatrix}\alpha&\beta\\-\beta&\alpha\end{bmatrix}
    \quad\text{and}\quad
    \begin{bmatrix}\alpha&\beta\\\beta&-\alpha\end{bmatrix},
    \qquad
    \alpha,\beta\in\mathbb{R},
\end{equation}
which form two 2-dimensional linear subspaces $\mathcal{S}_{+}$ and $\mathcal{S}_{-}$ of $\mathbb{R}^{2\times2}$.
The two subspaces are mutually orthogonal and together span $\mathbb{R}^{2\times2}$.
Orthogonal projection of $C$ gives
\begin{align}
    \operatorname{dist}^2(C,\mathcal{S}_{+})
    &=\tfrac{1}{2}(c_{11}-c_{22})^2+\tfrac{1}{2}(c_{12}+c_{21})^2,\\
    \operatorname{dist}^2(C,\mathcal{S}_{-})
    &=\tfrac{1}{2}(c_{11}+c_{22})^2+\tfrac{1}{2}(c_{12}-c_{21})^2,
\end{align}
and the nearer subspace is selected by $\eta=\operatorname{sign}(\det C)$, since
$\operatorname{dist}^2(C,\mathcal{S}_{-})-\operatorname{dist}^2(C,\mathcal{S}_{+})=2\det C$.

\subsection{Exact Equivalence of the Residuals}
\label{sec:suppl_equivalence}

Combining \cref{sec:suppl_isometry,sec:suppl_nullspace,sec:suppl_o2},
\begin{equation}
    \operatorname{dist}^2(A,\mathcal{E})
    =\tfrac{1}{2}(c_{11}-\eta c_{22})^2+\tfrac{1}{2}(c_{12}+\eta c_{21})^2,
\end{equation}
and dividing by $\|A\|_F^2=\|C\|_F^2$ shows that the two signed components of the residuals in \cref{eq:two_by_two_residual} satisfy $\|r\|_2^2=\ell(A)$ exactly.

\subsection{A Globally Smooth Orthonormal Completion}
\label{sec:suppl_basis}

This section verifies the properties of the completion \cref{eq:complement} of \cref{sec:smooth_evaluation}.
Write $w=(a_1-ua_3,\,a_2-va_3,\,0)^\top$ and
\begin{equation}
    n=\frac{w\times e_3}{\|w\times e_3\|}
     =\frac{1}{\|w\|}\bigl(a_2-va_3,\ -(a_1-ua_3),\ 0\bigr)^\top,
\end{equation}
neither of which depends on the focal length.
By \cref{eq:null_vector_plane}, $\hat a\in\operatorname{span}\{w,e_3\}$ for every $f>0$, and $n$ is orthogonal to both $w$ and $e_3$, so $n^\top\hat a=0$ on the entire domain.
Both are unit vectors, so $\|\hat a\times n\|=\|\hat a\|\,\|n\|=1$ and $\hat a\times n$ is orthogonal to both.
The rows of \cref{eq:complement} are therefore orthonormal and span $\hat a^\perp$ without any renormalization, and $P_a$ is as smooth as $\hat a$ itself, that is, smooth for all $f>0$.
Since $n\times(\hat a\times n)=\hat a$, the frame $(n,\hat a\times n,\hat a)$ is right-handed for every $f$, so the basis also has constant orientation.
The same construction applied to $b$ and $K_i$ gives $P_b$.

This construction becomes degenerate in the special case where $w=0$, which corresponds to the epipole lying exactly at the principal point.
To handle this we can fix a coordinate axis not parallel to $\hat a$ and obtain the complement by Gram–Schmidt; since $(\hat a=\pm e_3)$ for all $f>0$, this fallback is smooth and well conditioned.
In practice we did not encounter any issues with this.

If we want to optimize the principal points as well, then $w$ will depend on these and the plane is no longer fixed.
In this case we can evaluate $n$ once at the initial principal point and complete $\hat a$ by Gram--Schmidt orthogonalization against it, followed by normalization.
The axis is exactly orthogonal to $\hat a$ at the initialization and the principal-point updates are small relative to the focal length, so the completion again remains well conditioned.

\subsection{The Determinant Branch Cannot Change}
\label{sec:suppl_branch}

By \cref{sec:suppl_isometry}, $|\det C|=\sigma_1\sigma_2>0$ whenever $A$ has rank two.
The calibration matrices $K$ are invertible for all positive focal lengths (and for any principal points), so $A_{ij}$ has rank two on the entire parameter domain.
The domain is connected and $\det C$ is a continuous function of the parameters, so its sign is constant on the domain.
This means that the branch $\eta$ (which is fixed at initialization) remains the nearest branch throughout optimization.

\section{Focal Length Initialization}
The initialization uses the focal-length prior stored for each image by COLMAP, together with the fundamental matrices of the incident view-graph edges.
When no manually specified or EXIF focal length is available, COLMAP sets this prior to $1.2\max(w,h)$.
For each edge, we compute the closed-form Bougnoux two-view focal estimates~\cite{bougnoux1998projective} and use them as votes for the two incident images.
The procedure is as follows.

\begin{enumerate}
    \item For each edge, shift the fundamental matrix according to the principal points of its two images and normalize it using the corresponding focal-length priors.
    \item Solve the two Kruppa constraints induced by the normalized fundamental matrix.
    The positive solutions give candidate focal lengths for both images.
    \item Discard candidates whose focal lengths are outside the range $[0.1, 10]$ times the corresponding COLMAP priors.
    If multiple candidates remain, retain the one with the smallest essential-manifold residual.
    \item Add the selected focal estimate as a vote for each incident image.
    \item For each image with at least three votes, accept the median vote when its interquartile spread is sufficiently small, specifically when $Q_{75}/Q_{25}
     < 1.2$.

    \item If an image has too few votes or inconsistent votes, scale its COLMAP prior by the scene-level median of the estimated-to-prior focal ratios over all valid votes.
    If no valid votes are available, this ratio is set to one and the original prior is retained.
\end{enumerate}

The resulting focal lengths initialize the subsequent robust view-graph refinement, while the principal points are kept fixed.

\end{document}